\documentclass{article}

\usepackage{PRIMEarxiv}

\usepackage[utf8]{inputenc} 
\usepackage[T1]{fontenc}    
\usepackage{hyperref}       
\usepackage{url}            
\usepackage{booktabs}       
\usepackage{amsfonts}       
\usepackage{nicefrac}       
\usepackage{microtype}      
\usepackage{lipsum}
\usepackage{tablefootnote}
\usepackage{color, colortbl}
\usepackage[misc]{ifsym} 
\usepackage{soul}
\usepackage{amsmath}
\usepackage{fixltx2e}
\usepackage{afterpage}
\usepackage{caption}
\usepackage{fancyhdr}       
\usepackage{graphicx}       
\graphicspath{{media/}}     
\usepackage[sorting=none, style=numeric-comp]{biblatex} 
\definecolor{brown}{RGB}{229,225,224}
\usepackage{float}

\usepackage{graphicx}
\usepackage[flushleft]{threeparttable}

\title{Artifact Reduction in Undersampled 3D Cone-Beam CTs using a Hybrid 2D-3D CNN Framework

}

\author{
  Johannes Thalhammer$^\mathrm{{1-4}}$ \Letter, \hspace{0.3em} Tina Dorosti$^\mathrm{{1-3}}$, Sebastian Peterhansl$^\mathrm{{1, 2}}$, Daniela Pfeiffer$^\mathrm{{3, 4}}$,\\
  \textbf{Franz Pfeiffer$^\mathrm{{1-4}}$, Florian Schaff$^\mathrm{{1, 2}}$} \\\\
  1 Chair of Biomedical Physics, Department of Physics, School of Natural Sciences\\
  2 Munich Institute of Biomedical Engineering\\
  3 Institute for Diagnostic and Interventional Radiology, School of Medicine and Health, TUM Klinikum\\
  4 Institute for Advanced Study\\\\
  Technical University of Munich, Germany\\
  \Letter \hspace{0.3em} \texttt{johannes.thalhammer@tum.de} \\
}

\begin{document}
\maketitle

\begin{abstract}
Undersampled CT volumes minimize acquisition time and radiation exposure but introduce artifacts degrading image quality and diagnostic utility. Reducing these artifacts is critical for high-quality imaging. We propose a computationally efficient hybrid deep-learning framework that combines the strengths of 2D and 3D models. First, a 2D U-Net operates on individual slices of undersampled CT volumes to extract feature maps. These slice-wise feature maps are then stacked across the volume and used as input to a 3D decoder, which utilizes contextual information across slices to predict an artifact-free 3D CT volume. The proposed two-stage approach balances the computational efficiency of 2D processing with the volumetric consistency provided by 3D modeling. The results show substantial improvements in inter-slice consistency in coronal and sagittal direction with low computational overhead. This hybrid framework presents a robust and efficient solution for high-quality 3D CT image post-processing.\\
The code of this project can be found on github: \url{https://github.com/J-3TO/2D-3DCNN_sparseview/}
\end{abstract}
\section{Introduction}
Computed tomography (CT) has proven to be a versatile imaging modality with widespread applications in medical diagnostics, industrial inspection, and material science \cite{DeChiffre2014IndustrialTomography} \cite{Vasarhelyi2020MicrocomputedReview}. CT enables critical insights into complex systems by providing high-resolution cross-sectional views of internal structures. However, achieving high-quality CT volumes typically requires prolonged scanning times or elevated radiation doses, which can pose risks to patients, increase costs, or limit applicability in dynamic scenarios requiring high temporal resolution. Undersampled CT techniques address these concerns by reducing acquisition times and radiation exposure at the cost of introducing severe artifacts that compromise image quality and diagnostic utility.

To mitigate these artifacts, various techniques have been developed, ranging from analytical image processing to advanced deep learning methods \cite{Cheslerean-Boghiu2023WNet:Layer} \cite{Wu2021DRONE:Reconstruction} \cite{Hu2017AnReconstruction}. Although deep learning approaches have shown considerable promise, many primarily operate on 2D axial slices simulated with simplified scan setups such as parallel or fan beam geometries, disregarding the inherently volumetric nature of CT data, which is especially prominent for real data measured with a cone beam geometry. This often leads to inter-slice inconsistencies and limits the reconstruction quality. Fully 3D deep learning models address this limitation by leveraging volumetric context; however, their computational demands restrict model size, reducing their capacity to model complex contexts and effectively capture long-range dependencies critical for artifact reduction \cite{Sidorenko2021DeepSamples} \cite{Wang2021Multi-viewTransformers}.

To address these challenges, we propose a novel hybrid framework that synergistically combines 2D and 3D modeling. Our approach employs a 2D U-Net to efficiently extract features from individual slices and a 3D decoder to integrate these features for volumetric reconstruction. This hybrid architecture leverages the computational efficiency of 2D processing while preserving the volumetric consistency and structural fidelity of 3D modeling. We demonstrate that our method effectively reduces undersampling artifacts, achieving high-quality, artifact-free CT reconstructions with reduced computational overhead.

\section{Materials and Methods}
\subsection{Overview}
\begin{figure*}[!h]
  \centering
  \includegraphics[width=1.0\textwidth]{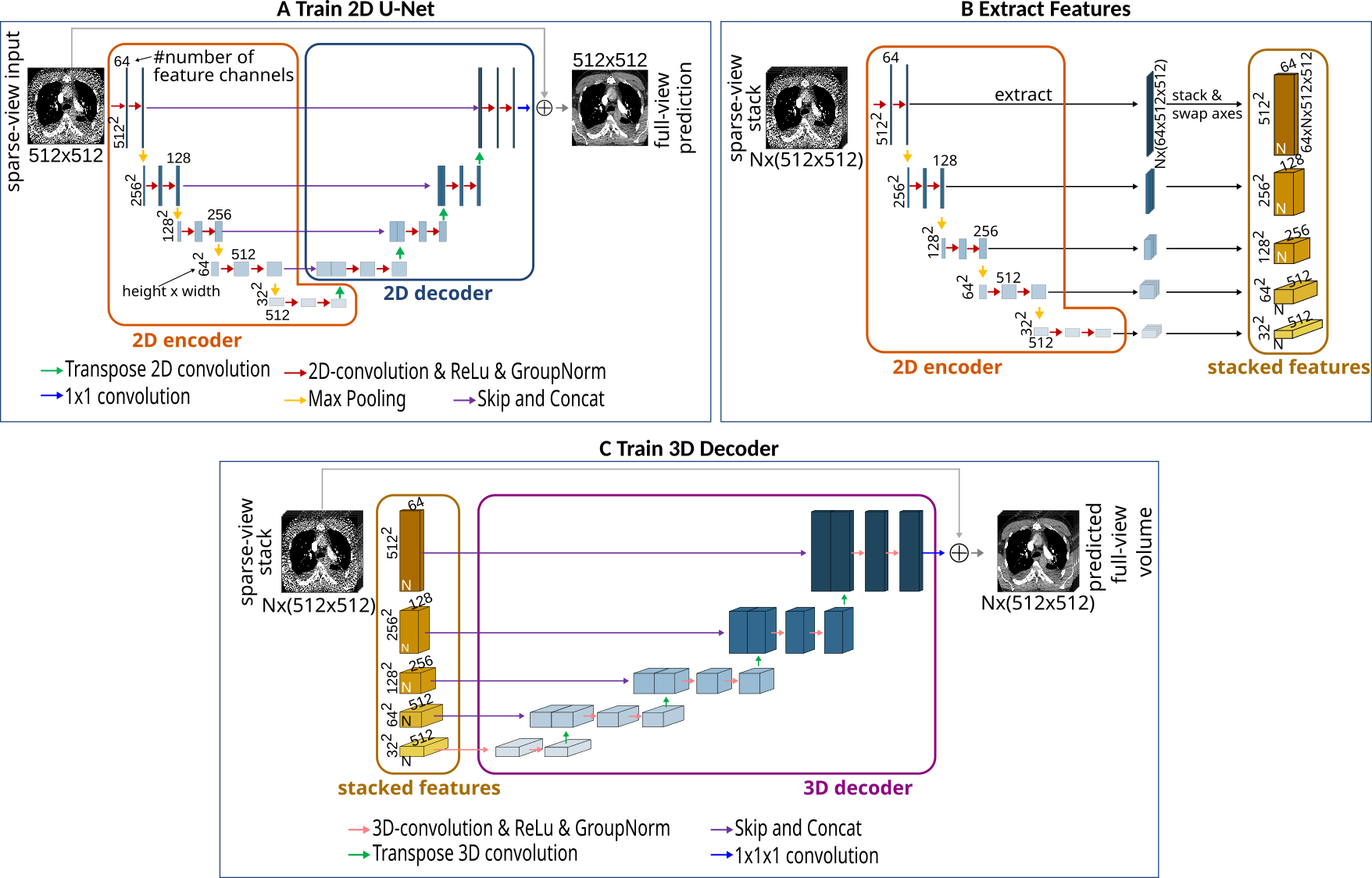}
  \caption{Overview of the 2D-3D training pipeline. A) First, a 2D U-Net is trained on 2D axial slices to remove artifacts. B) The encoder of the trained U-Net is used to extract features at different levels of an entire CT volume by looping through $N$ axial slices. Subsequently, the extracted features from a volume are stacked together, effectively transforming the 2D feature maps into 3D feature maps. C) The extracted 3D feature maps are used as input along with the sparse volume to train a 3D decoder, which returns an artifact-reduced CT volume.}
  \label{fig:methodoverview}
\end{figure*}

The proposed method is illustrated in Figure \ref{fig:methodoverview}. First, a 2D U-Net is trained on 2D axial slices to remove artifacts, using sparse-view images as input and full-view images as target (A) \cite{Ronneberger2015U-net:Segmentation}. The 2D U-Net comprises an encoding path with five convolutional blocks, connected by four pooling layers leading to a bottleneck resolution of 32x32 for a 512x512 input. This is followed by a decoder that progressively upsamples the resolution again.
In stage (B), the trained encoder of the 2D U-Net is utilized to extract features from an entire CT volume by sequentially processing $N$ axial slices. The obtained 2D feature maps are then stacked, resulting in a 3D feature representation of the volume.
In the final stage (C), the stacked 3D feature maps, along with the sparse CT volume, are used as input to a 3D decoder. The 3D decoder follows the architectural principles of the 2D decoder but employs 3D convolutions with a (3x3x3) kernel. This volumetric processing enables the integration of contextual information across slices, producing an artifact-reduced 3D CT volume with improved inter-slice consistency.

\subsection{Dataset}
The dataset used in this study was obtained from the RSNA Pulmonary Embolism Detection Challenge (2020), comprising 7,279 CT pulmonary angiograms \cite{Colak2021TheDataset}. Sparse-view reconstructions were generated by forward-projecting the data using a cone-beam geometry with only 128 views, followed by reconstruction using FDK, implemented by the ASTRA toolbox \cite{vanAarle2016FastToolbox}. For testing, 100 CT scans were set aside and excluded from all training processes. The 2D U-Net was trained using a dataset comprising 5,251 scans for training and 1,313 scans for validation, randomly sampled from the remaining data. For training the 3D decoder, data from 150 scans were used for training, with 50 scans reserved for validation.

\subsection{Training Details}
The 2D U-Net was trained on 512x512 axial slices using an NVIDIA RTX 3090 GPU. Features were subsequently extracted, stacked, and saved. The 3D decoder was then trained on an NVIDIA A100 GPU using the extracted features from 48 randomly selected neighboring slices during training. All training was conducted using PyTorch and PyTorch Lightning using MSE loss and the AdamW optimizer \cite{Paszke2019PyTorch:Library}. Training continued until validation loss converged, and the models with the lowest validation loss were selected.

\section{Results}
\begin{figure*}[h]
  \centering
  \includegraphics[width=1.0\textwidth]{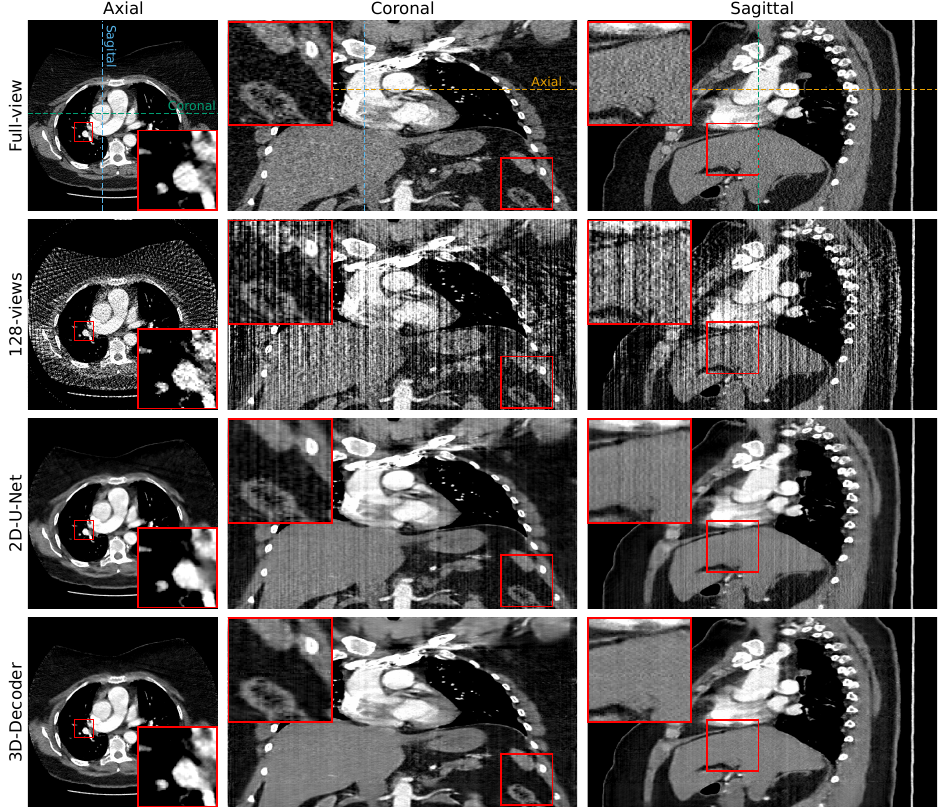}
  \caption{An axial, coronal, and sagittal slice of the full-view reconstruction from the test split (first row), 128-view reconstruction (second row) with artifact reduction by the 2D U-Net (third row) and by the Hybrid 2D-3D CNN framework (last row), respectively. Images are in the mediastinum window (Width:350 HU, Level:50 HU). The dotted lines indicate the location of the different views. All inserts are 75x75 pixels.}
  \label{fig:Results}
\end{figure*}

Figure \ref{fig:Results} presents a comparison of artifact reduction performance between the 2D U-Net and the 3D decoder. In the axial plane, both methods substantially reduce undersampling artifacts, yielding visually comparable results. However, in the coronal and sagittal views, differences become evident. The 2D U-Net leaves noticeable artifacts in the vertical direction, compromising volumetric consistency. In contrast, the 3D decoder effectively mitigates these artifacts, achieving superior artifact reduction in these planes and improving the overall volumetric quality.

Table \ref{tab:metrics} presents a quantitative evaluation of the methods, featuring PSNR and SSIM values computed from a test dataset that consists of 100 volumes, each with an average of 244 slices. Both, the 2D U-Net and the 3D decoder substantially improve these metrics compared to the sparse-view reconstructions. The 2D U-Net achieves slightly higher PSNR and SSIM values.  In terms of processing time, the 2D U-Net post-processing achieves an average runtime of 2.408$\pm$0.651 seconds per CT volume. By comparison, the 3D decoder, including the feature extraction step, requires 20.276$\pm$5.768 seconds per volume, both measured on an NVIDIA A100 GPU.

\begin{table}
  \centering
  \begin{tabular}{lcc}
  \toprule
    & PSNR & SSIM \\
  \midrule
  sparse-view       & 24.810 $\pm\,$ 0.522      &   0.637  $\pm\,$ 0.015     \\
  2D U-Net          & 39.289 $\pm\,$ 1.295      &   0.949  $\pm\,$ 0.015   \\
  3D decoder        & 38.086 $\pm\,$ 1.188     &   0.938  $\pm\,$ 0.016   \\

  \bottomrule
  \end{tabular}
  \caption{SSIM and PSNR values from the sparse-view data without post-processing, with 2D U-Net post-processing and with 3D decoder post-processing, respectively. The values were calculated on the test split (100 CT scans).}
  \label{tab:metrics}
\end{table}

\section{Discussion}

The visual results demonstrate that the 2D-3D CNN framework effectively combines the advantages of axial 2D training and volumetric 3D decoding. The 2D U-Net focuses on reducing undersampling artifacts primarily present in the axial plane, while the 3D decoder improves interslice consistency. Interestingly, the metrics for the purely 2D approach are slightly better. This discrepancy is likely due to the smaller training set used for the 3D decoder, which was constrained by the significant storage requirements for the extracted features. It is anticipated that extending the training set—possibly through on-the-fly feature extraction during training and leveraging multiple GPUs—will further enhance the 3D decoder's performance. Additionally, the effectiveness of this approach should be evaluated on downstream tasks such as disease detection or segmentation to better understand the practical benefits of this framework.

\section{Conclusion}
This exploratory study demonstrates that a hybrid 2D-3D CNN framework, which combines slice-wise 2D feature extraction with volumetric 3D decoding, effectively reduces artifacts in undersampled cone-beam CT volumes. The approach not only achieves good inter-slice consistency but also maintains a low computational overhead, making it a promising approach for improving image quality.


\newpage
\printbibliography 
\end{document}